\documentclass[conference]{IEEEtran}
\IEEEoverridecommandlockouts

\usepackage{cite}
\usepackage{amsmath,amssymb,amsfonts}
\usepackage{algorithmic}
\usepackage{graphicx}
\usepackage{textcomp}
\usepackage{xcolor}
\usepackage{booktabs}
\usepackage{tabularx}
\usepackage{enumitem}
\usepackage{url}

\usepackage{fancyhdr}
\usepackage{etoolbox}

\fancypagestyle{firstpage}{
    \fancyhf{}
    \fancyhead[C]{\small A PREPRINT}
    \fancyfoot[C]{\small This work is a preprint and has not been peer-reviewed. It has been posted to share preliminary findings. The views expressed here are those of the authors and do not necessarily reflect the views or policies of ISRO SAC.  }

}

\begin{document}

\title{Evaluation of image simulation open source solutions for simulation of synthetic images in lunar environment}

\author{\IEEEauthorblockN{1\textsuperscript{st} Jai G Singla}
\IEEEauthorblockA{\textit{Space Applications centre} \\
\textit{Indian Space Research Organization}\\
Ahmedabad, India \\
jaisingla@sac.isro.gov.in}
\and
\IEEEauthorblockN{2\textsuperscript{nd} Hinal B Patel }
\IEEEauthorblockA{\textit{hinalbpatel@gmail.com}}
\and
\IEEEauthorblockN{3\textsuperscript{rd} Nitant Dube}
\IEEEauthorblockA{\textit{Space Applications centre} \\
\textit{Indian Space Research Organization}\\
Ahmedabad, India \\
nitant@sac.isro.gov.in}
}

\maketitle
\thispagestyle{firstpage}

\begin{abstract}
Synthetic image generation is one of the crucial input for planetary missions. It enables
researchers and engineers to visualize planned planetary missions, test imaging systems and
plan exploration activities in a virtual environment before actual deployment. Image
simulation is essential for assessing landing sites, detecting hazards, and validating
navigation systems in a missions. This study offers a detailed evaluation of various image
simulation approaches for the lunar environment, with particular emphasis on the effects of
different camera models and light illumination conditions on the quality of synthetic lunar
images. These images are produced using real Digital Elevation Models (DEM) and terrain
data derived from instruments such as Chandrayaan-2 Orbiter High Resolution Camera
(OHRC) and NASA’s Wide Angle Camera (WAC), and Narrow Angle Camera (NAC)
instruments. This research aims to improve the reliability of synthetic imagery in supporting
autonomous navigation and decision-making systems in lunar exploration. This work
contributes to the development of more effective tools for generating important information
for future lunar missions and enhances the understanding of the moon’s surface environment.
\end{abstract}


\section{Introduction}
The development of an image simulation software is crucial for the development of various
fields such as computer vision, robotics, remote sensing and virtual reality, and it plays a
particularly vital role in space exploration. For lunar missions, simulation tools enable testing
and evaluation of spacecraft sensors, rovers and remote sensing instruments without relying
solely on costly space missions or ground testing. Simulating the lunar synthetic images
effectively prepares systems for lunar missions, ensuring reliability and success. The
appearance of the lunar surface is strongly influenced by both camera characteristics and
solar illumination geometry. Variation in the sun angle affect shadow formation and feature
visibility, especially in challenging areas such as the lunar South Pole, where the low angle of
sunlight results in extended shadow and harsh lighting. Variations in sensor resolution, field
of view and viewing geometry significantly affect the quality and characteristics of imagery
captured by different instruments. Therefore, excessive reliance on real datasets limits the
ability to generalize across various mission scenarios and environment conditions. To
overcome these limitations, image simulation techniques have been developed to replicate
realistic lunar surface appearance by integrating terrain models, illumination conditions, and
sensor characteristics.

Early simulation approaches relied on simplified shading models and low-resolution terrain
data, which restricted their accuracy. However, recent advancements have significantly
improved simulation fidelity through the use of high-resolution Digital Elevation Models
(DEMs), physically based reflectance models, and precise ephemeris data for accurate
positioning of the Sun and spacecraft. Specialized tools such as PANGU \cite{parks2004}
and SurRender \cite{leb2021} are commercial software tools used in professional
space applications, providing high quality image simulation based on physical models. These
tools enable accurate modeling of shadowing effects, illumination geometry, and sensor
characteristics, including noise and optical distortions. In contrast, general-purpose platforms
like Blender\cite{blender} and Unity offer flexible environments for visualization and rapid
prototyping, although they often require additional customization to achieve scientific
accuracy. Quantum Geographic Information System (QGIS) \cite{qgis} is an open source
geographic information system commonly used in remote sensing for analyzing and
visualizing geospatial data, including DEM alignment, map projection, and data preparation.
It supports various remote sensing data formats. Furthermore, simulation frameworks like A
physically based rendering application in MATLAB (ABRAM) \cite{abram} and The
Celestial Object Rendering Tool (CORTO) \cite{corto} facilitate the integration of
trajectory data, camera models, and illumination conditions, enabling the generation of time-
varying image sequences for mission-specific analysis. CORTO, a python library built on
Blender, is used to model space imaging scenarios with high realism. The framework
provides large synthetic images with depth maps. These simulated outputs are also validated
against real mission data to ensure accuracy and reliability. ABRAM is an evolving
MATLAB-based simulation framework. It generates realistic synthetic images of celestial
bodies such as planets, moons, asteroids and deep space environments by incorporating
physical models of terrain, light and atmospheric conditions. It supports the design and
testing of optical and imaging systems through the simulation of diverse scenarios, including
surface landing, orbital observation and satellite missions. Additionally, it simulates the effect
of space weather, sensor noise and illumination variations, providing crucial data for mission
planning, scientific analysis and the development of algorithms for navigation, mapping and
object recognition. Table \ref{tab:comparision_table} summarizes the key differences between various simulated
software.

\begin{table*}[t]
\centering
\scriptsize
\caption{Comparison of simulation software}
\label{tab:comparision_table}
\begin{tabularx}{\textwidth}{XXXXXX}
\toprule
Feature & PANGU & SurRender & Blender & CORTO & ABRAM \\
\midrule

Type & Space-focused & Space-focused & General 3D &  Research Framework & Research Framework \\
Realism & High &  Very High & Medium-High & High & High \\
Physics-based Rendering & Yes & Yes (advanced) & Limited & Yes & Yes \\
DEM Support & Yes &  Yes &  Yes & Yes & Yes \\
Camera Modeling &  Accurate & Very Accurate & Custom & Accurate & Accurate \\
Illumination modeling & Strong &Very strong & Moderate & Strong & Strong \\
Ease of Use & Moderate & Complex & Easy &  Easy & Easy \\
Speed & Medium & Optimized  & Fast & Medium & Medium \\
Best Use Case & Space Simulation & High-fidelity rendering & Visualization & Simulation & Space Simulation \\
Software type & Commercial & Commercial & Open source & Open source & Open source \\
\bottomrule
\end{tabularx}
\end{table*}

\section{Dataset Details}

Lunar observation missions have generated extensive datasets with varying spatial resolutions
and coverage. These datasets enhance the scientific understanding of the Moon surface. The
lunar datasets considered in this study are:
\begin{enumerate}[label=(\roman*)]
\item Chandrayaan 2 OHRC: The Chandryaan-2 Orbiter High Resolution Camera (OHRC)
\cite{ohrc}, operated by Indian Space Research Organization (ISRO), provides
sub-meter panchromatic imagery of the lunar surface with a resolution of 30 cm. Its
high spatial resolution enables detailed mapping of small-scale terrain features, such
as boulders, craters and ridges.
\item LROC NAC: The LROC NAC \cite{nac}, operated by NASA, captures high-
resolution panchromatic images of the moon with a spatial resolution of 0.5-2 meters.
It provides stereo image pairs that enable precise generation of DEMs through
photogrammetry. NAC gives detailed, local information that complements broader
datasets.
\item LROC WAC: The LROC WAC \cite{wac}, a NASA instrument, provides near-
global coverage at 100 m/pixel in monochrome and multispectral bands. WAC images
help to study the Moon’s surface at a large scale, showing regional terrain and
lighting conditions. They are useful for understanding the overall geological context
and for planning safe landing sites.
\item LOLA DEM: The LOLA \cite{loladem} provides high-precision topographic
measurements of the Moon’s surface. LOLA uses laser pulses to measure surface
elevations with a vertical accuracy of approx. 1 meter and a horizontal sampling
resolution ranging from 25 to 50 meters. The LOLA DEMs capture detailed large and
regional-scale topography, including crates, ridges and slope variations.
\end{enumerate}

\section{Tools and Technologies}
MATLAB R2022a \cite{matlab} along with Image processing Toolbox, Parallel
Computing Toolbox and Statistics and Machine learning Toolbox was used to generate
synthetic lunar images. Additionally, Blender \cite{blender} and Python \cite{python} were
utilized to simulate and render realistic lunar scenes. Data pre-processing was carried out
using QGIS software \cite{qgis} and python programming.

\section{Methodology}
The methodology of lunar image simulation involves integrating real datasets, terrain models,
illumination conditions and camera parameters to generate realistic synthetic images. The
specific approaches and tools used for this process are detailed in this section.

\begin{figure}[h]
\centering
\includegraphics[width=\linewidth]{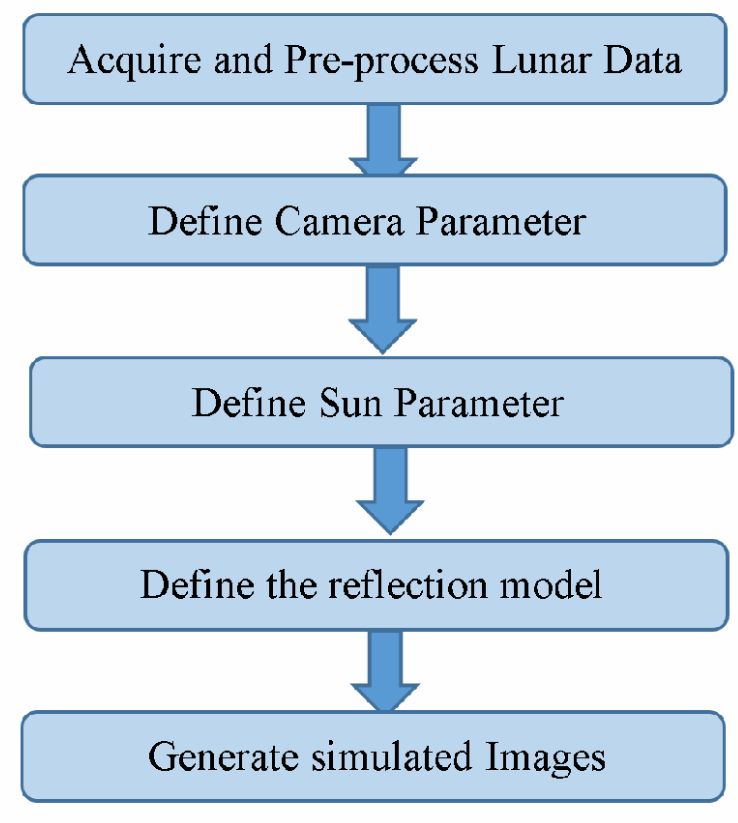}
\caption{Flow Diagram}
\label{fig:abram_flow}
\end{figure}

System level diagram to generate synthetic images using ABRAM is presented in Figure \ref{fig:abram_flow}.The process involves selecting input data such as DEM and reference lunar images from sensors like OHRC, NAC, WAC and LOLA DEM, which are then preprocessed for proper
alignment, scaling and regional selection. Subsequently, the simulation environment is
configured by specifying lighting parameters such as light source type, temperature and
radius. And the textures maps include Albedo, Displacement, Normal and Horizon maps.
Albedo map defines the surface’s reflectivity. Local height difference relative to reference
surface is encoded in displacement maps. Normal map indicates the surface’s local
orientation enabling effective modeling of shadows caused by viewing conditions. And
horizon map defines area affected by the planet’s curvature. Together, these textures enable a
detailed and realistic visualization and analysis of the lunar surface. Furthermore, realistic
light interaction is achieved through the selection of appropriate radiometric models such as
Lambertian or Hapke reflectance, which define how light is reflected by the surface.\cite{villa2023} Lambertian reflection is a basic and widely used model that assumes surfaces scatter incident light equally in all directions, resulting in a matte appearance. Lambertian model is defined as equation \ref{eq:1}.In contrast, the Lommel-seeliger model expressed in equation \ref{eq:2}, provides more sophisticated representation of surface reflectance, especially for bodies with irregular or rough surfaces. The Hapke model is used to describe how light interacts with rough or uneven surfaces. It considers factors like single-scattering albedo, phase function and surface roughness to explain the bidirectional reflectance, including effects like multiple scattering and shadowing as mention in equation \ref{eq:3}. This radiometric models makes it useful for accurately modeling the brightness and appearances of planetary surfaces under varying lighting conditions. The camera model is configured by specifying parameters such as focal length, exposure time, and camera orientation and position. Focal length determines the zoom level of a camera by controlling the magnification and the field of view (FOV). A longer focal length makes things appear bigger and closer, while a shorter focal length offers
wider field of view (FOV). Camera orientation is described by roll, pitch and yaw, representing the rotation angles around the camera’s longitudinal, lateral, and vertical axes respectively. The scene geometry is then defined through simplified parameters such as distance, phase angle and orientation angles. Once all parameters are defined, the framework performs the rendering process by simulating interaction between light and surface to generate synthetic images under different viewing and illumination conditions.

\begin{equation}
\label{eq:1}
I = I_0 \cos \theta
\end{equation}

\begin{equation}
\label{eq:2}
I = \frac{I_0}{\pi} \cdot \frac{1}{r^2}
\end{equation}

\begin{equation}
\label{eq:3}
I = I_0 \left( \frac{\cos \theta_1 \, \cos \theta_2}{\pi} \right) \left( \frac{1 - B_0}{1 + B_0} + B_0 \right)
\end{equation}
Where, $I_0$ is the incident light intensity. $\theta$ is the angle between the light direction and surface normal. $r$ is the distance between the light source and the surface. $\theta_1$ is the incident angle and is $\theta_2$ the observation angle. $B_0$ is the backscatter parameter.

\begin{figure}[h]
\centering
\includegraphics[width=\linewidth]{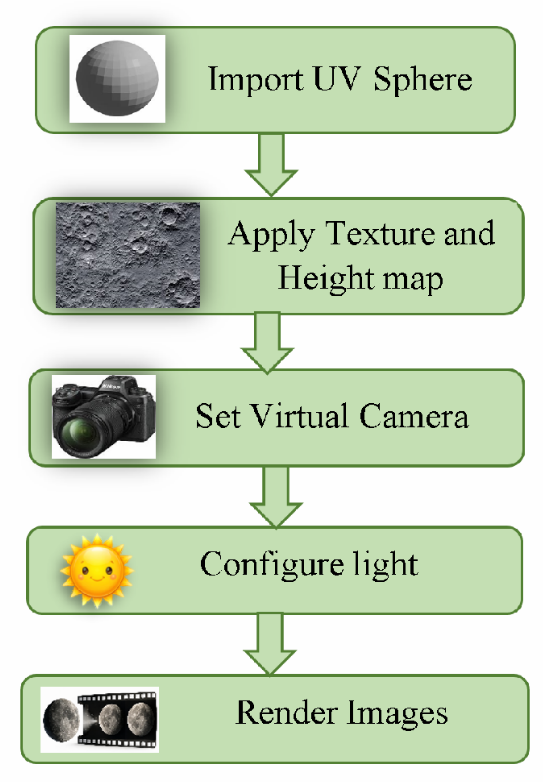}
\caption{Flow Diagram of Image Simulation using Blender}
\label{fig:blender_flow}
\end{figure}

In Blender, generating a realistic lunar image involves serval key steps as illustrated in Figure \ref{fig:blender_flow}.Lunar image generation starts by importing a high-resolution 3D model of the moon. Surface details such as craters and terrain are enhanced using textures and normal maps to add realism. The scene setup includes positioning the camera to simulate desired view and
configuring lighting to mimic sunlight, often adjusting the phase angle to create realistic
shadows. Rendering is performed using ray tracing engines to accurately simulate light
interactions with the surface. . Finally, sequences of images can be generated by rotating the
Moon or camera to create animations for mission planning, scientific visualization or
illumination studies.

The moon surface visualization in QGIS 3D was achieved by importing high-resolution
terrain data and lunar elevation data. These datasets were then accurately aligned to create a
realistic terrain model. The 3D map view was activated and then elevation data was
configured as the terrain height field. Additional adjustments, such as vertical exaggeration,lighting and shading effects were applied to enhance surface details. Finally, navigation tools
were used to explore the lunar landscape from various perspectives, resulting in an immersive
and detailed 3D visualization for analysis.

Lastly, an in-house SW was developed to simulate and model the camera's perspective in a
3D environment. The simulation takes pitch, roll, and yaw angles as inputs to define the
camera's orientation. Camera height is also provided as an input parameter, representing the
vertical position of the camera in the scene. Using these parameters, the code calculates the
camera's viewing point and line of sight. It computes the resolution for each pixel based on
the camera's intrinsic properties and field of view. The simulation incorporates a base texture
map to provide surface details for the environment. Additionally, a height map is used to
simulate terrain elevation and surface variations. These maps help generate a realistic scene
by adding depth and texture to the environment. The combined data allows for rendering the
scene as viewed from the specified camera position and orientation. This setup can be useful
for visualizations, simulations, or virtual environment rendering.

\section{Results and Discussions}
This section presents the synthetic images simulated using various open source software’s
considering Lunar environment such as sun elevation and azimuth angle, height and viewing
geometry of the acquisition sensor and terrain characteristics .

\begin{figure}[h]
\centering
\includegraphics[width=\linewidth]{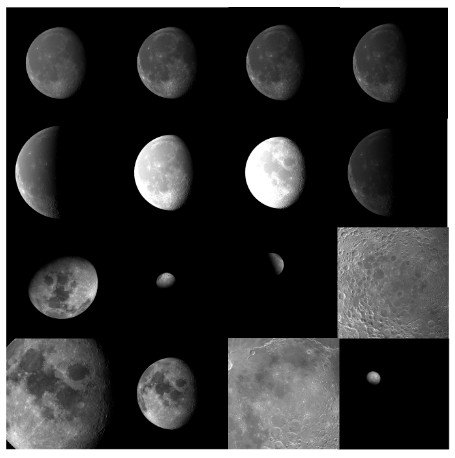}
\caption{Moon Simulated images under different sun angle and phase angle}
\label{fig:1}
\end{figure}

Figure \ref{fig:1} illustrates simulation of synthetic images over moon under different lighting
conditions and phase angles. It utilizes target input of WAC lunar mosaic \cite{wac}, which
has a resolution of 100 meters per pixels. The simulation also used the LOLA DEM to
incorporate terrain elevation data. This visualization illustrates generation of synthetic images
based on appearance of Moon under different lighting condition and different phase angle.
The phase angle is important because it influences the visibility of surface features. It also
affects the scattering of light, thereby affecting the observed characteristics of the Moon. This
analysis provides valuable insights into how surface features and topography of moon change
with varying environmental parameters.

\begin{figure}[h]
\centering
\includegraphics[width=\linewidth]{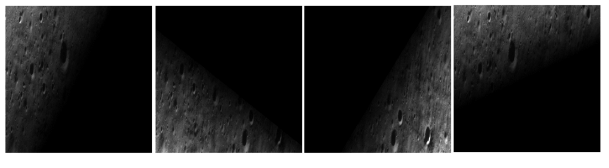}
\caption{OHRC Data Visualization under different sun angles}
\label{fig:2}
\end{figure}

Figure \ref{fig:2} depicts the OHRC \cite{ohrc} simulated data from different sun angles. The
variation in illumination significantly affects the visibility of surface features. When the sun
is at different positions details of designated areas become clearer, while vice versa for other
areas. These dynamic shift in illumination helps reveal different parts of the terrain that might
be hidden or less clear with specific light angle. Using images from multiple sun angles,
provides a fuller understanding of the surface, making it an important method for accurately
studying lunar surface data.

\begin{figure}[h]
\centering
\includegraphics[width=\linewidth]{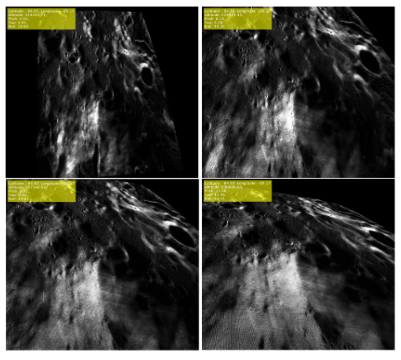}
\caption{NAC data visualization under various altitude}
\label{fig:3}
\end{figure}

Figure \ref{fig:3} presents the NAC data along with LOLA DEM over South Pole of the moon at
different altitude and camera angles. These datasets are valuable for generating realistic
landing simulations. Simulation of images from different altitudes and camera angles allows
for better assessment of terrain features and obstacles from multiple perspectives. This helps
improve the planning and safety of landing procedures. Using this data ensures more accurate
and reliable landing scenarios. Figure 5 simulate NAC data at latitude -84.83º and longitude -
85.17º. In Figure \ref{fig:3}(a), depicts the simulated data at altitude of 514285 meter with orientation angle of 4.91º pitch, 4.91º yaw and 19.64 º Roll. Similarly, Figure \ref{fig:3}(b) shows reproduced data under 228571-meter altitude with 8.91º pitch, 8.91º yaw and 35.20 º Roll. Figure \ref{fig:3}(c) displays modeled data at 157142-meter altitude with camera orientation of 9.82º pitch, 9.82º yaw and 44.61 º Roll. Lastly, Figure 5(d) illustrates generated data with 100000-meter altitude and 11.46º pitch, 11.46º yaw and 54.43 º Roll.

\begin{figure}[h]
\centering
\includegraphics[width=\linewidth]{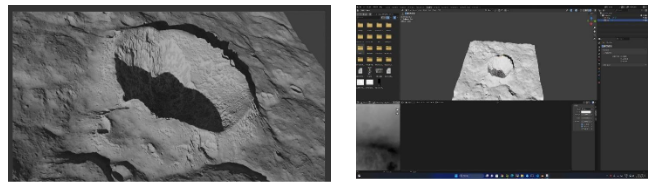}
\caption{Moon simulation in Blender}
\label{fig:4}
\end{figure}

Figure \ref{fig:4} represents simulated moon image using blender. Blender produces realistic 3D
models with customizable lighting and camera settings. It allows users to enhance terrain
features with textures and detailed height details, improving the overall visualization quality.
However, processing large datasets may require significant computational resources. Blender
does not support full GIS functionality and should be used alongside specialized GIS
software.

\begin{figure}[h]
\centering
\includegraphics[width=\linewidth]{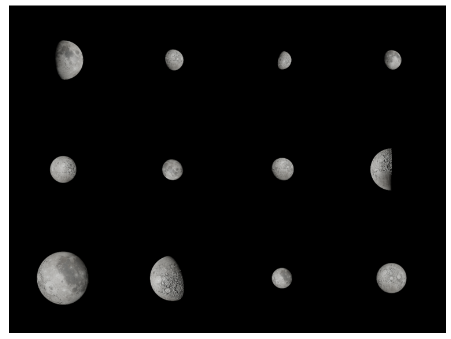}
\caption{Moon simulated images using CORTO}
\label{fig:5}
\end{figure}

\begin{figure}[h]
\centering
\includegraphics[width=\linewidth]{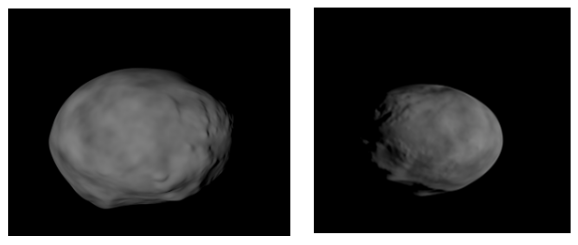}
\caption{Didymos asteroid simulated by CORTO}
\label{fig:6}
\end{figure}

Simulated Moon images and didymos asteroid with different angles, generated by CORTO,
are presented in Figure \ref{fig:5} and \ref{fig:6} respectively. These images showcase the Moon and didymos asteroid from various perspectives, highlighting its surface features and topography. The use of CORTO allows for accurate and diverse visualizations, which are essential for detailed
analysis. These simulations can be useful for training models, planning lunar missions, or
conducting scientific research.

\begin{figure}[h]
\centering
\includegraphics[width=\linewidth]{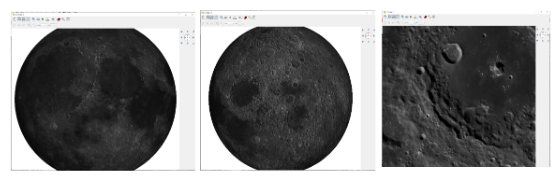}
\caption{3D View in QGIS}
\label{fig:7}
\end{figure}

\begin{figure}[h]
\centering
\includegraphics[width=\linewidth]{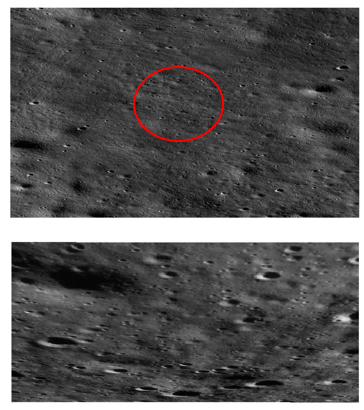}
\caption{Simulated image using Python (A) Input image (B) Simulated image}
\label{fig:8}
\end{figure}

Figure \ref{fig:7} displays a 3D view of lunar surface created using QGIS software. This view presents immersive view of lunar surface. This perspective helps in examining elevation and terrain
feature more effectively. Figure 10 presents a simulation of the lunar data viewed from a
specified camera perspective using python. Figure \ref{fig:8} (A) illustrate input OHRC data at -
69.37 latitude and 32.34 longitude. In Figure \ref{fig:8} (B), simulated image with camera orientation set to a roll angle 0.5, pitch angle of -40.7 and camera height of 0.1 kilometre is presented.

In this study, we evaluated various lunar simulation approaches including ABRAM, CORTO,
Blender, QGIS and python based rendering. ABRAM is a MATLAB based simulation
software whereas CORTO is python based simulation software. ABRAM and CORTO both
software effectively simulates images of various space object with different camera and sun
position and orientation. Both software are validated against real-world scenarios to assess
their accuracy. Blender offers effective shadow effects and high quality rendering but
extensive customization of camera and light positions is complicated. Additionally, Blender
depends on GIS software for date pre-processing and struggles with handling large datasets.
This limitation makes this software less suitable for large-scale space simulation scenarios.
On the other hand, QGIS 3D view has great 3D visualization capabilities and provides tilling
option at multiple zooming levels. However, it is inaccurate in simulating shadows and
camera perspectives. Our python based simulation provides a flexible and computationally
efficient method of simulating the lunar surface from a camera perspective. This approach
has also been validated against real scenarios, although it currently does not support detailed
lighting and shadow effects.

\section{Conclusions}
The rendering frameworks presented in this study provides a robust and physically consistent
approach for simulating images of lunar surface. By integrating accurate terrain data,
elevation data, realistic illumination based on solar positioning and precise camera
simulation, it enables the generation of synthetic images which closely resemble real mission
data. This capability is particularly valuable for research and development as it supports the
testing and validation of navigation algorithms, hazard detection techniques, and landing
strategies in a controlled environment. Furthermore, the flexibility of the framework allows
researchers to model diverse scenarios, such as varying illumination conditions at the lunar
poles. The generated synthetic images can also be utilized to train deep learning models for
various tasks such as hazard detection and landing site selection, thereby enhancing the
robustness and reliability of autonomous systems. Overall, tools evaluated in this study serves
as an effective tool for advancing planetary exploration studies by bridging the gap between
theoretical modeling and real-world mission requirements. Out of all the evaluated tools, ABRAM provides all the simulated scenarios required for the simulation of realistic synthetic
images.

\end{document}